\documentclass[sigconf,nonacm]{acmart}
\usepackage{enumitem}

\setcopyright{none} % 移除版权声明
\AtBeginDocument{%
  }

\renewcommand\footnotetextcopyrightpermission[1]{}

\acmConference{}{}{}
\acmBooktitle{}

\title{BLURR: A Boosted Low-Resource Inference for Vision-Language-Action Model}

\author{
\textbf{
Xiaoyu Ma$^{\ast,\dagger}$,
Zhengqing Yuan$^{1,\ast}$, \\
Zheyuan Zhang$^{1}$,
Kaiwen Shi$^{1}$,
Lichao Sun$^{2}$,
Yanfang Ye$^{1}$
}\\[6pt]
$^1$University of Notre Dame, \quad
$^2$Lehigh University
}

\begin{document}

\begin{abstract}
Vision--Language--Action (VLA) models enable impressive zero-shot
manipulation, but their inference stacks are often too heavy for
responsive web demos or high-frequency robot control on commodity GPUs.
We present \textbf{BLURR}, a lightweight inference wrapper that plugs
into existing VLA controllers without retraining.
Instantiated on $\pi_0$, BLURR keeps the original checkpoints and
observation interfaces, and accelerates control by combining an
instruction-prefix KV cache, mixed-precision execution, and a
single-step rollout schedule.
Our demo lets attendees switch between controllers and toggle
inference options in real time while watching SimplerEnv episodes.
We will release BLURR and integration scripts as open source at
\url{https://github.com/JijiKing-Sam/BLURR-A-Boosted-Low-Resource-Inference-for-Vision-Language-Action-Model}.
\end{abstract}


\maketitle

% 强制样式调整（在 maketitle 之后）
\pagestyle{plain}
\thispagestyle{plain}  
\renewcommand{\thefootnote}{\fnsymbol{footnote}}
\footnotetext[1]{Equal contribution}
\footnotetext[2]{Xiaoyu Ma is an independent researcher student, remotely working with Yanfang Ye.}

\section{Introduction}

Large, generalist policies pretrained on diverse robot data, such as Octo~\cite{octo}, OpenVLA~\cite{openvla}, and Pi-0~\cite{pi0}, have recently made Vision-Language-Action (VLA) control an accessible abstraction for real-world manipulation.
By conditioning on natural-language instructions and multi-view images, these models can generate joint-space or end-effector actions and adapt to new embodiments with modest in-domain data.
However, despite their impressive capability, the heavy vision encoders (e.g., SigLIP) and multimodal decoders (e.g., PaliGemma-style backbones) impose substantial computational overhead: per-step inference can exceed 30–50,Hz latency requirements, especially when hundreds of visual tokens and long language prefixes are processed at every control cycle~\cite{flashattention,openvla}.
This gap between model capacity and system responsiveness creates a practical barrier for interactive deployment, e.g., browser demos or real-time teleoperation, where slow control loops can degrade user experience and even compromise task feasibility.
As a result, there is a growing need for methods that retain the strengths of existing VLA checkpoints while improving inference-time efficiency to support reliable, real-time robotic control.
\begin{figure}[t]
  \centering
  \includegraphics[width=\linewidth]{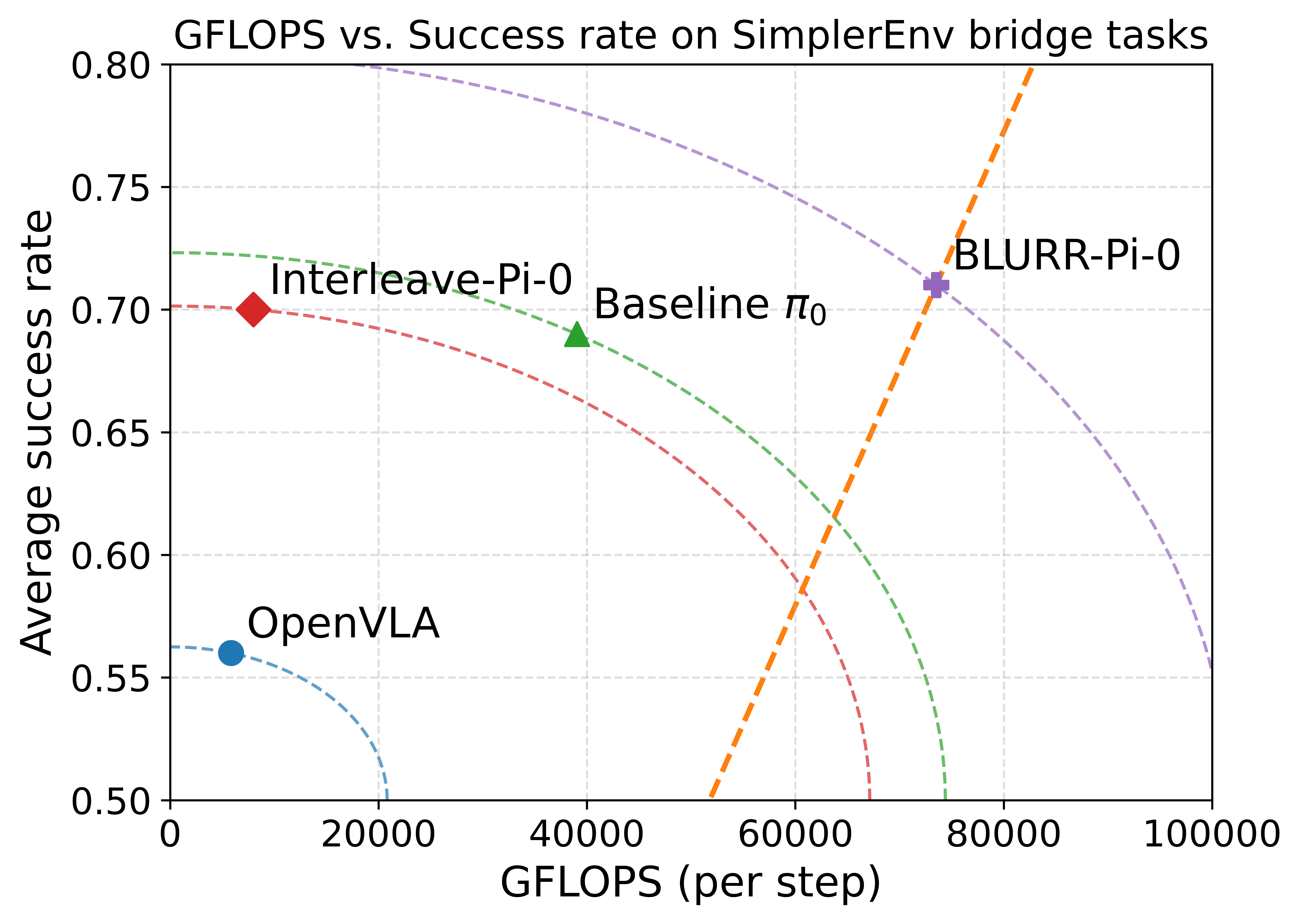}%
  \caption{
    Efficiency--performance landscape on four SimplerEnv bridge tasks.
    Each marker shows a controller's per-step compute throughput (GFLOPS, $x$-axis)
    and average task success rate (0--1, $y$-axis).
    The dashed circular guides are centered at the origin and pass through each
    model, so that larger radii qualitatively indicate a more extreme operating
    regime in the (compute, success) plane.The orange dashed ray is the line through the origin and \textbf{BLURR-Pi-0}, highlighting that our inference wrapper moves along a high-efficiency
    direction.
}
    \vspace{-15pt}
  \label{fig:gflops-success}
\end{figure}

Several recent works attempt to address this latency issue, but existing approaches face notable limitations.
1) First, many efficiency-oriented methods require substantial changes to the underlying architecture or training pipeline, for example, TinyVLA~\cite{tinyvla} redesigns the backbone and action head for compactness, and MiniVLA \cite{minivla} compresses OpenVLA through structural modifications.
Such interventions force practitioners to retrain models from scratch, often demanding hundreds of GPU-hours and risking performance regressions on downstream tasks.
2) Second, token-reduction or action-reparameterization techniques (e.g., FAST~\cite{fast}) introduce new tokenization schemes that are incompatible with released checkpoints, preventing them from serving as drop-in optimizers for widely used VLA models.
3) Third, methods that improve reasoning or spatial-temporal grounding—such as CoA-VLA’s chain-of-affordance~\cite{coavla} or TraceVLA’s visual-trace prompting~\cite{tracevla}, enhance capability but do not meaningfully reduce the per-step compute cost in interactive control loops.
Consequently, none of these techniques directly provide a lightweight inference acceleration layer that preserves existing model checkpoints while achieving real-time responsiveness.
Therefore, there is a clear need for a technique that \textbf{\textit{preserves existing VLA checkpoints without retraining, delivers substantial reductions in per-step inference latency, and maintains the original policy’s accuracy and manipulation performance.}}

\begin{figure}[t]
  \centering
  \includegraphics[width=\linewidth]{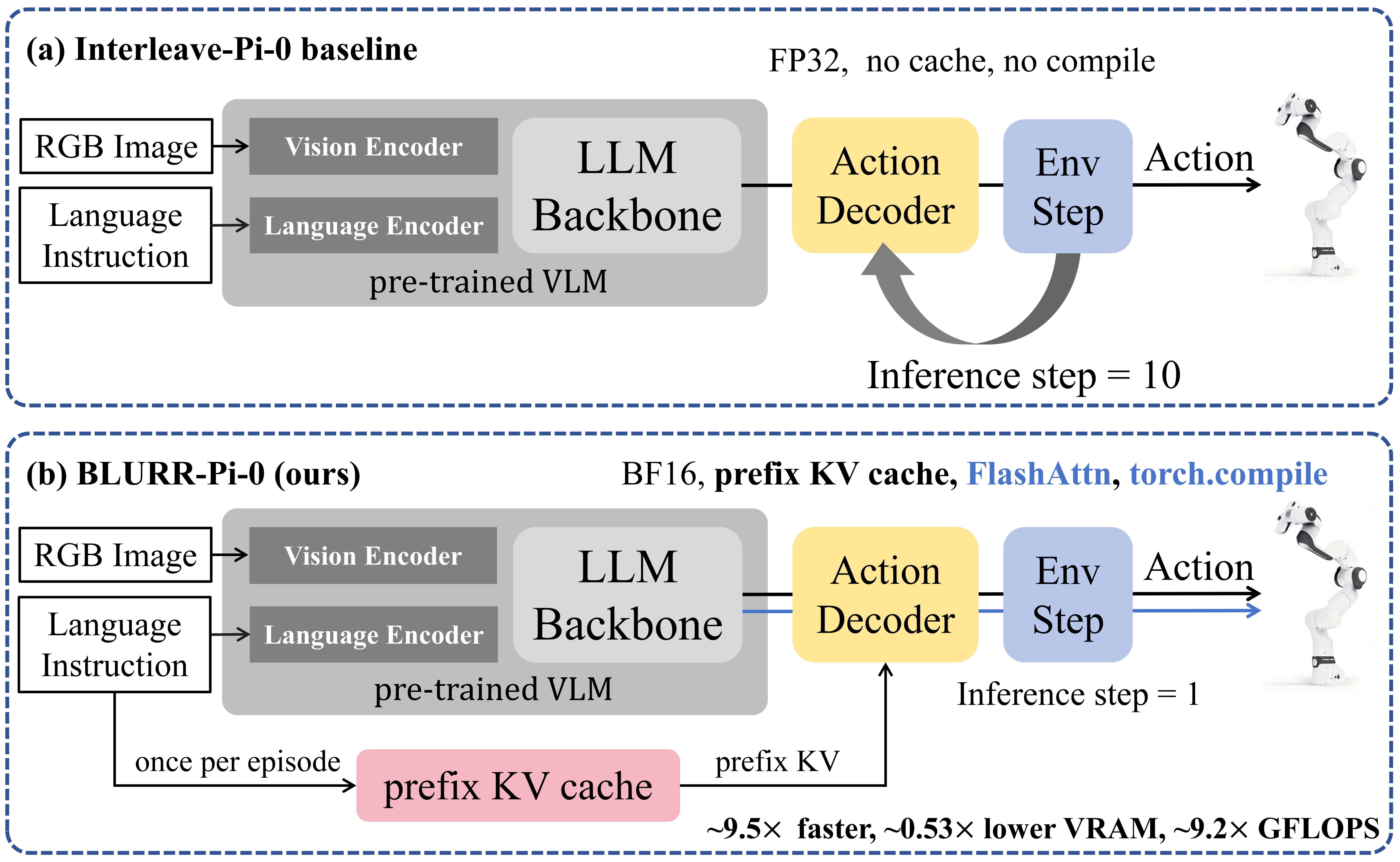}
  \caption{High-level comparison of the Interleave-Pi-0~\cite{interleavevla} baseline and our BLURR-Pi-0 inference stack.
In (a), RGB images and language instructions are jointly encoded by a pre-trained vision--language model and decoded into actions over a 10-step control rollout using an FP32 action decoder without caching or compilation.
In (b), BLURR-Pi-0 keeps the same pre-trained VLM backbone but moves instruction processing into a one-time prefix KV cache, runs the action decoder in BF16 with FlashAttention and \texttt{torch.compile}, and reduces the control horizon to a single inference step.}
    \vspace{-15pt}
  \label{fig:overview}
\end{figure}
To address these limitations, we introduce BLURR, a lightweight inference wrapper designed to accelerate VLA controllers while preserving their original checkpoints.
Instead of modifying network weights or redesigning tokenization schemes, BLURR focuses solely on the inference pipeline, reorganizing how visual and language tokens are processed during control.
1) First, BLURR requires no architectural changes and no retraining: any existing OpenVLA-, Pi-0–, or TraceVLA-style checkpoint can be wrapped without altering model parameters or task logic.
2) Second, BLURR substantially reduces per-step latency through a combination of BF16 execution, compiled decoder graphs, prefix KV caching for instructions, and a single-step control horizon.
These inference-side optimizations collectively yield order-of-magnitude speedups while lowering memory consumption.
3) Third, despite its aggressive acceleration, BLURR maintains the original policy’s accuracy and manipulation performance, enabling real-time, browser-based VLA interaction without sacrificing task success.
In practice, BLURR preserves SOTA manipulation performance while delivering up to \textbf{9.5$\times$ lower latency}, \textbf{0.53$\times$ peak VRAM}, and \textbf{9.2$\times$ higher effective GFLOPS}, setting a new efficiency bar for VLA inference.
% \section{Background and Related Work}

% \subsection{Generalist VLA Models}

% Octo~\cite{octo} and OpenVLA~\cite{openvla} are representative VLA policies trained on hundreds of thousands of manipulation trajectories from Open X-Embodiment.
Pi-0~\cite{pi0} proposes a flow-matching action head for more dexterous, high-frequency control.
TraceVLA~\cite{tracevla} augments OpenVLA with visual trace prompting to improve spatial-temporal awareness in both simulation and real-robot experiments.
% Vision backbones such as SigLIP~\cite{siglip} and multimodal decoders such as PaliGemma~\cite{paligemma} support these models by providing strong visual-language representations, but they also introduce substantial compute and memory cost per frame.
% \subsection{Efficiency-Oriented VLAs}

% Several recent works explicitly tackle VLA efficiency.
MiniVLA~\cite{minivla} shrinks the OpenVLA backbone while preserving performance on LIBERO benchmarks.
TinyVLA~\cite{tinyvla} targets fast and data-efficient VLAs via compact backbones and diffusion decoders.
FAST~\cite{fast} focuses on efficient action tokenization, and CoA-VLA~\cite{coavla} improves generalization via a visual-text chain-of-affordance.
% Orthogonal to these architecture and training improvements, our work looks at the \emph{inference layer} for existing checkpoints: how to pool visual tokens, reuse pre-computed prefixes, and wire in FlashAttention~\cite{flashattention} without breaking API contracts for OpenVLA, Pi-0, or TraceVLA-style controllers.
% \subsection{Evaluation Environments}

% SimplerEnv~\cite{simplerenv} provides a suite of simulation environments designed to mirror real robot setups with matched sensors and task distributions.
It has quickly become a standard testbed for VLA evaluation, including TraceVLA~\cite{tracevla} and several follow-up works.
We adopt SimplerEnv for our demo because it lets us measure both success rate and system-level latency while keeping the control loop realistic.
\section{BLURR Architecture}

Our goal is to accelerate existing Vision–Language–Action (VLA) models at inference time without modifying their parameters, training procedure, or API.
BLURR keeps all model weights unchanged, but restructures the computational pathway to eliminate redundant work, minimize memory traffic, and exploit hardware-efficient kernels.
As shown in Figure 1, our method is built around three principles: 1) reduce redundant prefix computation, 2) minimize per-step token cost, and 3) maximize tensor-core utilization.
\subsection{Single-step control with prefix-cached instructions}
\label{subsec:prefix-kv}

BLURR keeps the underlying controller's $224{\times}224$ vision encoder
and token budget, and instead reduces per-step compute by reusing
instruction tokens across the whole episode.
Let $c$ be the language instruction and
$o_t = (I_t, s_t)$ the observation at time $t$ (RGB image $I_t$ and
state $s_t$).
We write
\begin{equation}
  P = T(c) \in \mathbb{R}^{L_p \times d},\qquad
  V_t = E_v(I_t, s_t) \in \mathbb{R}^{L_v \times d},
\end{equation}
where $T$ and $E_v$ are the frozen text and vision--state encoders.
The input sequence to the Transformer backbone at step $t$ is
\begin{equation}
  X_t = [P; V_t] \in \mathbb{R}^{(L_p+L_v)\times d}.
\end{equation}
\noindent\textbf{During control.}
A standard Interleave-Pi-0 controller recomputes keys and values for
all $(L_p{+}L_v)$ tokens at every step.
BLURR-Pi-0 instead caches the instruction prefix once per episode:
\begin{equation}
  K_{\text{pref}}^{(\ell)} = P W_K^{(\ell)},\quad
  V_{\text{pref}}^{(\ell)} = P W_V^{(\ell)},
\end{equation}
and at step $t$ only projects the visual--state tokens
\begin{equation}
  K_{\text{step},t}^{(\ell)} = V_t W_K^{(\ell)},\quad
  V_{\text{step},t}^{(\ell)} = V_t W_V^{(\ell)},
\end{equation}
forming
\begin{equation}
  K_t^{(\ell)} =
  \begin{bmatrix}
    K_{\text{pref}}^{(\ell)}\\
    K_{\text{step},t}^{(\ell)}
  \end{bmatrix},
  \quad
  V_t^{(\ell)} =
  \begin{bmatrix}
    V_{\text{pref}}^{(\ell)}\\
    V_{\text{step},t}^{(\ell)}
  \end{bmatrix}.
\end{equation}
With BLURR's single-step rollout (one forward pass per environment
step), the instruction cost is paid once per episode instead of at every
control step, yielding the per-step latency reductions reported in
Table~\ref{tab:latency}.
\begin{table}[t]
   \centering
   \caption{Single-step efficiency comparison across the original Pi-0 controller, our Interleave-Pi-0 baseline, and the proposed BLURR-Pi-0 variant, using the same input resolution and token budget (224$\times$224 RGB, 256 tokens) on single H100 GPU.
BLURR nearly doubles the effective GFLOPS compared to the Pi-0 baseline.}
   \label{tab:latency1}
   \begin{tabular}{lcccc}
     \toprule
     Configuration & Latency (ms) & VRAM (GB) & GFLOPS\\
     \midrule
     OpenVLA            & 217.8 & 14.33 & 5{,}835 \\
     OpenVLA-OFT        & 91.2 & 14.48 & 49{,}886 \\
     Pi-0 baseline      & 111.6 & 13.58 & 39{,}038 \\
 
     Interleave-Pi-0    & 162.1 & 13.61 &  7{,}989 \\
     \textbf{BLURR-Pi-0(ours)} & \textbf{17.1} & \textbf{7.20} & \textbf{73{,}525} \\
     \bottomrule
   \end{tabular}
 \end{table}

\subsection{Efficient BF16 Decoder with Compilation and FlashAttention}

The action decoder remains the main performance bottleneck, and BLURR accelerates it through three inference-time techniques:

\noindent\textbf{1) BF16 execution.}  
All decoder layers are executed in BF16, which reduces memory bandwidth requirements by roughly $2{\times}$ and enables full utilization of contemporary tensor-core accelerators (e.g., H100).
Weights remain unchanged; only runtime casting is used.

\noindent\textbf{2) Compiled computation graph.}  
We wrap the entire forward pass in a \texttt{torch.compile} graph, enabling kernel fusion and eliminating Python overhead.
This converts the decoder into a streamlined compute pipeline with significantly fewer dispatch points.

\noindent\textbf{3) FlashAttention kernels.}  
When attention shapes allow, BLURR enables fused, IO-aware attention kernels via the PyTorch SDPA backend.
FlashAttention greatly reduces memory I/O during multi-head attention, improving both latency and VRAM efficiency.
\noindent Together, these optimizations convert the decoder into a high-throughput inference engine capable of sub-20\,ms per-step latency on a single modern GPU, \emph{without altering the model’s parameters}.
% \noindent\textbf{Drop-in and configurable.}
% The wrapper should present the same observation and action interfaces as an existing Interleave-VLA controller such as Interleave-Pi-0, so that research repos can adopt it by swapping a small amount of inference code rather than retraining models or changing task logic.
At the same time, all major inference knobs are exposed as configuration flags (e.g., \texttt{use\_bf16}, \texttt{use\_compile}, \texttt{num\_inference\_steps}), so that users can directly see how each choice affects latency, memory, and task success.
% \noindent\textbf{Low-latency control.}
% The full perception--language--action loop should run at interactive frequencies on a single modern GPU, so that a WWW attendee can issue natural-language commands and immediately see the robot respond.
Concretely, we target sub-20\,ms per control step on an H100 GPU, corresponding to $\sim$50--60\,Hz control.
% \noindent\textbf{High-throughput efficiency.}
% Beyond wall-clock latency, the inference stack should make good use of available hardware, turning FLOPs into useful control updates rather than Python overhead or redundant computation.
We therefore track approximate per-step GFLOPS and peak VRAM for each configuration, and design BLURR to substantially increase effective throughput while basically keeping the underlying $\pi_0$ checkpoint fixed.
% \subsection{From Interleave-\texorpdfstring{$\pi_0$}{pi0} to BLURR-\texorpdfstring{$\pi_0$}{pi0}}

% Figure~\ref{fig:overview} illustrates how BLURR wraps an existing Interleave-Pi-0 controller.
In the baseline design (Figure~\ref{fig:overview}a), RGB observations and language instructions are concatenated and fed through a pre-trained vision--language backbone, after which an VLA decoder produces a sequence of actions over a 10-step control rollout in FP32.
Every control step re-encodes the full instruction prompt, runs the decoder in eager mode without KV caching, and executes on standard attention kernels.
This yields a responsive but relatively heavy inference stack: on our H100 setup the Interleave-Pi-0 baseline requires roughly 162\,ms per step and 13.6\,GB of peak VRAM.
% BLURR keeps the \emph{same} $\pi_0$ checkpoint and observation pipeline, but replaces the inference stack with a lightweight controller (Figure~\ref{fig:overview}b).
The wrapped controller, which we denote BLURR-$\pi_0$, runs the decoder in BF16, uses a compiled forward graph, and reduces the control horizon to a single step.
Instructions are processed once at the beginning of an episode to build a prefix KV cache, and subsequent control steps only inject fresh visual and proprioceptive tokens.
Under this configuration we measure roughly 17\,ms per step and 7.2\,GB of peak VRAM on the same H100 GPU, corresponding to about $9.5\times$ lower latency and $0.53\times$ peak memory while retaining the original $\pi_0$ weights.
% \subsection{Inference Stack Optimizations}

% \noindent\textbf{Single-step control horizon.}
% Interleave-style controllers commonly unroll a short-horizon sequence of actions such as 10 steps for every high-level decision.
For short tabletop manipulation tasks in SimplerEnv, we found that this horizon is over-provisioned: a single-step controller can achieve comparable success rates while drastically reducing compute.
BLURR therefore sets inference steps to 1 and delegates temporal smoothing to the underlying environment and low-level controllers.
This immediately removes a $10\times$ factor in wall-clock latency for many tasks.
% \medskip
% \noindent\textbf{BF16 and compiled decoder.}
% BLURR runs the VLA decoder in BF16 to exploit H100 Tensor Cores and cut memory traffic roughly in half, without changing model weights or fine-tuning.
We further wrap the full image--text--action forward pass in \texttt{torch.compile} using a reduce-overhead mode, which fuses kernels and reduces Python dispatch overhead.
Where available, we enable efficient attention kernels such as FlashAttention~\cite{flashattention} through PyTorch's SDPA backend.
Together, BF16 + compilation turn the decoder into a high-throughput inference engine that delivers the measured $9.5\times$ latency and $9.2\times$ GFLOPS improvements over the FP32 eager baseline.
% \medskip
% \noindent\textbf{Prefix KV cache for instructions.}
% In the Interleave-Pi-0 baseline, the full language instruction is concatenated with visual tokens at every control step, so the decoder repeatedly re-processes the same prompt.
BLURR isolates the instruction into a \emph{prefix} segment that is encoded once per episode.
We run a short forward pass to build a KV cache for these prompt tokens, then reuse this cache at every subsequent control step by appending fresh visual and proprioceptive tokens on top.
This design requires careful tokenizer alignment and attention-mask construction to keep positions consistent, but it amortizes instruction processing and makes per-step cost depend primarily on the number of visual tokens.
% \medskip
% \noindent\textbf{Drop-in integration.}
% All of the above changes are implemented as a thin wrapper around the original $\pi_0$ model class and its Hydra configuration files.
The same checkpoint can be evaluated as either Interleave-Pi-0 or BLURR-$\pi_0$ by flipping a small set of flags, and the demo UI simply exposes these flags as interactive toggles.
As a result, BLURR demonstrates that substantial gains in VLA responsiveness can be achieved by re-engineering the inference stack alone, without altering architectures or retraining models.
\begin{figure}[t]
  \centering
  \includegraphics[width=\linewidth]{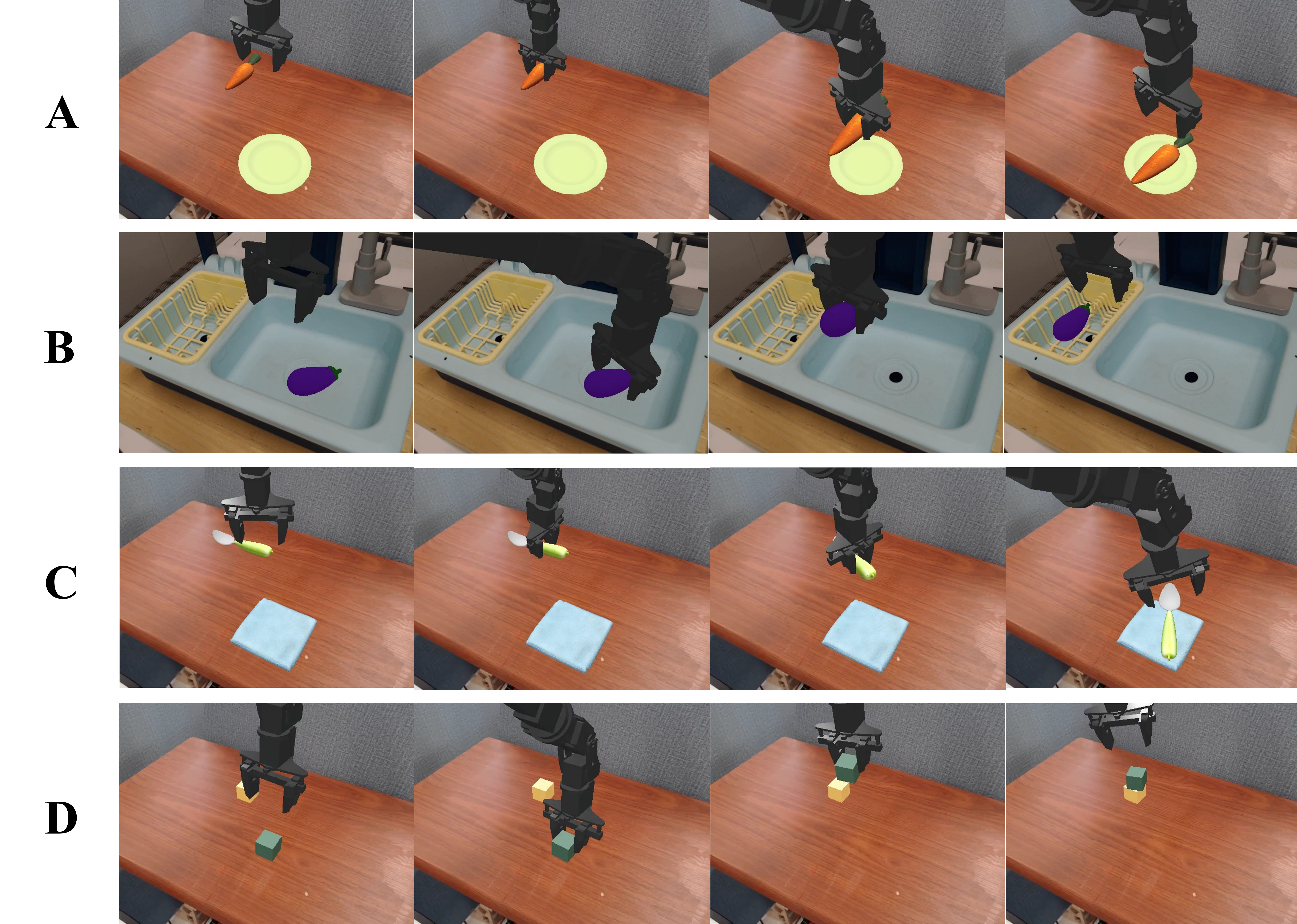}
  \caption{Example rollouts of BLURR-Pi-0 on the four SimplerEnv Bridge manipulation tasks.
Rows A--D show successful episodes for carrot-on-plate, eggplant-in-rack, spoon-on-cloth, and block-stacking respectively, with frames progressing from left to right in time.
Despite the aggressively accelerated inference stack, BLURR-Pi-0 produces smooth, goal-directed behaviours that closely match the intended task specifications.}
    \vspace{-15pt}
  \label{fig:ui}
\end{figure}

\section{Implementation}

\subsection{Runtime Stack and Demo Integration}

Our implementation is built on PyTorch and HuggingFace-style model hubs.
On the vision and VLA side we reuse: (i) OpenVLA checkpoints~\cite{openvla} implemented with PaliGemma-style multimodal decoders~\cite{paligemma}, and (ii) official reference implementations for Pi-0~\cite{pi0} and TraceVLA~\cite{tracevla}.
BLURR is implemented as a thin wrapper around these existing model classes and their Hydra configuration files.
The wrapper preserves the original observation and action interfaces of Interleave-style controllers, while exposing additional inference flags such as \texttt{use\_bf16}, \texttt{use\_compile}, and \texttt{num\_inference\_steps}.
At the system level, Figure~\ref{fig:overview} illustrates how BLURR intercepts the standard perception--language--action loop: camera observations and low-dimensional state are encoded by the frozen backbone, the instruction prompt is processed once per episode to build a prefix KV cache, and subsequent control steps only inject fresh visual and proprioceptive tokens.
Where available, we route attention through PyTorch's SDPA backend with FlashAttention kernels~\cite{flashattention};
otherwise we fall back to standard attention without changing model code.
For the WWW demo, we deploy this runtime inside a small client--server stack.
A backend process runs the VLA policy and SimplerEnv~\cite{simplerenv} instances on a single GPU.
A lightweight WebSocket bridge streams RGB frames, actions, and scalar metrics.
On the front end, the web UI in Figure~\ref{fig:ui} visualizes live trajectories and exposes BLURR's inference knobs (BF16, compilation, control horizon, KV caching) as interactive toggles.
Each toggle change triggers a short warm-up phase and then logs per-step latency, peak VRAM, and task outcomes, allowing attendees to directly observe the speed or accuracy trade-offs.
GFLOPS is the FLOP count reported by PyTorch’s profiler divided by per-step latency.
\subsection{Experimental Setup}

We evaluate BLURR on four in-domain SimplerEnv bridge tasks~\cite{simplerenv}: \emph{Carrot-on-plate}, \emph{Eggplant-in-container}, \emph{Spoon-on-plate}, and \emph{Stack-blocks}.
All experiments share the same Pi-0 checkpoint~\cite{pi0} and differ only in the control architecture and inference configuration.
We compare:
\begin{itemize}[left=0pt]
  \item \textbf{Pi-0 baseline}, which directly decodes actions from images and language without fine-tuning through interleaving inputs.
\item \textbf{Interleave-Pi-0}, our reproduction of an Interleave-VLA controller.
  \item \textbf{BLURR-Pi-0}, our accelerated variant.
\end{itemize}
Unless otherwise noted, we profile per-step latency, peak reserved GPU memory, and approximate GFLOPS on an NVIDIA H100 GPU by repeatedly running the same image--prompt pair through the controller.
For each configuration and task we then run 100 closed-loop evaluation episodes in SimplerEnv to measure task success rates.
\section{Evaluation}

\subsection{Overall Efficiency and Throughput}

Table~\ref{tab:latency1} summarizes the efficiency of our three controllers when evaluated with a fixed input resolution (224$\times$224 RGB) and token budget (256 tokens).
Despite sharing the same Pi-0 checkpoint, the three stacks exercise the hardware very differently:

Interleave-Pi-0 incurs additional decoding overhead from its 10-step rollout and eager FP32 execution, yielding higher latency and lower effective GFLOPS than the Pi-0 baseline.
In contrast, BLURR-Pi-0 combines a single-step controller, BF16 Tensor Core execution, and compiled graphs to reduce per-step latency by roughly $9.5\times$ and peak VRAM by about $0.53\times$ relative to Interleave-Pi-0, while nearly doubling the effective GFLOPS compared to the original Pi-0 controller.
\subsection{Ablation: BLURR Inference Components}

To isolate which parts of the BLURR stack matter most, we treat the ``Impact of BLURR components on per-step latency and VRAM`` table as a small ablation study.
Rather than changing network architecture or retraining, we progressively enable inference-side optimizations on top of the Interleave-Pi-0 baseline:

\begin{table}[t]
  \centering
  \caption{
    Impact of BLURR components on per-step latency and peak VRAM usage for a single Interleave-Pi-0 checkpoint evaluated on a SimplerEnv task~\cite{simplerenv} on an NVIDIA H100 GPU.
}
  \label{tab:latency}
  \begin{tabular}{lcc}
    \toprule
    Configuration & Latency (ms) & VRAM (GB) \\
    \midrule
    Interleave-Pi-0 (FP32, 10 steps)
      & 162.1 & 13.61 \\

    + BF16 only (10 steps)
      & 88.2 & 13.58 \\

    + \texttt{torch.compile} (10 steps)
      & 56.7 & 6.15 \\

    + fewer flow steps (6 steps)
      & 44.7 & 7.28 \\

    + fewer 
flow steps (4 steps)
      & 34.8 & 7.29 \\

    + KV cache
      & 31.9 & 7.32 \\

    + FlashAttention~\cite{flashattention}
      & 27.4 & 7.30 \\

    \midrule
    Full BLURR (ours, 1 step)
      & 17.1 & 7.20 \\

    \bottomrule
  \end{tabular}
\end{table}

In the current demo, the same ablations are exposed as toggles in the user interface, and Table~\ref{tab:latency} summarizes their cumulative effect on latency and memory. Conceptually, this breakdown matches our implementation: BF16 primarily reduces memory traffic, efficient attention and \texttt{torch.compile} reduce kernel launch and Python overhead, and the single-step controller with prefix KV caching removes redundant prompt processing.

\subsection{Closed-loop Success Rates and Demo Experience}

Finally, we assess whether these efficiency gains come at the cost of task performance.
Table~\ref{tab:bridge-success} reports success rates over 100 evaluation episodes for each of the four bridge tasks, comparing the original Pi-0 controller, our Interleave-Pi-0 baseline, and BLURR-pi-0:

\begin{table}[t]
  \centering
  \caption{Bridge task success rates for five controllers on four SimplerEnv in-domain tasks.
All entries are success probabilities (0--1) over 100 evaluation episodes per task.
}
  \label{tab:bridge-success}
  \begin{tabular}{lccccc}
    \toprule
    Model           & Carrot & Spoon & Blocks & Eggplant & Avg.
\\
    \midrule
    OpenVLA                     & 0.47 & 0.44 & 0.63 & 0.68 & 0.56 \\
    MiniVLA            & 0.42 & 0.67 & \textbf{0.69} & 0.18 & 0.49 \\
    Baseline $\pi_0$            & 0.53 & 0.84 & 0.53 & 0.88 & 0.69 \\
    Interleave-Pi-0    
 & \textbf{0.59} & 0.89 & 0.53 & 0.79 & 0.70 \\
    BLURR-Pi-0 (ours)           & 0.54 & \textbf{0.91} & 0.46 & \textbf{0.93} & \textbf{0.71} \\
    \bottomrule
  \end{tabular}
\end{table}

Across these four tasks, BLURR-Pi-0 matches or slightly exceeds the average success rate of Interleave-Pi-0 while delivering an order-of-magnitude reduction in per-step latency.
In the demo UI, attendees can switch between the three controllers and immediately observe the qualitative differences in control frequency: the BLURR-Pi-0 variant updates at roughly 50--60\,Hz and can react to sudden obstacles or perturbations much more quickly, while the Interleave-Pi-0 baseline produces visibly ``chunkier'' motion at around 6\,Hz.
This side-by-side comparison complements the quantitative results above and highlights BLURR's main message: substantial VLA speedups are achievable by re-engineering the inference stack alone, without retraining or modifying the underlying model checkpoints.

Our future work includes porting BLURR-style inference wrappers to other
VLA families and real robot platforms, beyond our current
SimplerEnv-based evaluation.
We also plan to move to longer-horizon, multi-stage manipulation tasks
where control frequency and latency may interact more strongly with task
difficulty.
Another direction is to explore adaptive scheduling and mixed-precision
policies that react to hardware load in real time, rather than using
fixed configurations.
Finally, we are interested in tighter integration with distillation or
compression techniques so that model design and inference wrappers can
be co-optimized instead of tuned in isolation.
% \section{Discussion and Limitations}

% Interleave-VLAcc is deliberately scoped: it does not attempt to redesign the underlying VLA architectures~\cite{minivla,tinyvla,fast,coavla}, nor does it tackle training-time efficiency.
Instead, it acts as a thin inference layer that can be combined with existing and future efficiency-oriented VLAs.
Our current implementation focuses on manipulation tasks in SimplerEnv~\cite{simplerenv}; extending it to mobile manipulation and long-horizon planning would require additional engineering and may interact with the assumptions made by Pi-0~\cite{pi0} and TraceVLA~\cite{tracevla}.
% \begin{acks}
% We thank the maintainers of OpenVLA, Octo, Pi-0, TraceVLA, SimplerEnv and FlashAttention for releasing their code and models, and colleagues for feedback on early versions of this demo.
% \end{acks}



\bibliographystyle{ACM-Reference-Format}
\bibliography{main}

\end{document}